\pdfoutput=1

\documentclass[11pt]{article}

\usepackage{ACL2023}

\usepackage{times}
\usepackage{latexsym}

\usepackage[T1]{fontenc}

\usepackage[utf8]{inputenc}

\usepackage{microtype}

\usepackage{inconsolata}

\usepackage[whole]{bxcjkjatype}
\usepackage{color,soul}
\setul{}{0.9pt} 
\definecolor{Zero}{rgb}{0.320,0.859,1}
\definecolor{Easy}{rgb}{0.158,0.625,0.565}
\definecolor{NotEasy}{rgb}{0.918,0.764,0.436}
\definecolor{Difficult}{rgb}{0.956,0.640,0.383}
\definecolor{HistogramOrange}{rgb}{1,0.5,0.055}
\definecolor{HistogramBlue}{rgb}{0.124,0.468,0.706}
\usepackage{capt-of}

\usepackage{paralist}

\usepackage{float}
\usepackage{placeins}

\usepackage{adjustbox}
\usepackage[capitalise,noabbrev]{cleveref}
\usepackage{booktabs}
\usepackage{array} 
\usepackage{multirow}
\usepackage{calc}	
\usepackage{graphicx}

\newcommand{\split}{%
    \textcolor{HistogramBlue}{%
        \nolinebreak\,\nolinebreak%
        \raisebox{1pt}[0pt]{$\big |$}%
        \,%
    }%
}

\title{Japanese Lexical Complexity for Non-Native Readers: A New Dataset}

\author{ \\
    \textbf{Yusuke Ide${}^{1}$ $\;\;\;$ Masato Mita${}^2$ $\;\;\;$ Adam Nohejl${}^1$ $\;\;\;$ Hiroki Ouchi${}^{1,3}$ $\;\;\;$ Taro Watanabe${}^1$} \\
    ${}^1$Nara Institute of Science and Technology \quad
    ${}^2$CyberAgent Inc. \quad
    ${}^3$RIKEN
    \\
    \texttt{\{ide.yusuke.ja6, nohejl.adam.mt3, hiroki.ouchi, taro\}@is.naist.jp,} \\
    \texttt{mita\_masato@cyberagent.co.jp} \\
}

\begin{document}


\maketitle


\begin{abstract}

Lexical complexity prediction (LCP) is the task of predicting the complexity of words in a text on a continuous scale. 
It plays a vital role in simplifying or annotating complex words to assist readers.
To study lexical complexity in Japanese, we construct the first Japanese LCP dataset.
Our dataset provides separate complexity scores for Chinese/Korean annotators and others to address the readers' L1-specific needs.
In the baseline experiment, we demonstrate the effectiveness of a BERT-based system for Japanese LCP.

\end{abstract}


\section{Introduction}

Reading comprehension requires a certain level of vocabulary knowledge.
The results reported by \citet{Hu2000} suggest that most English learners need to understand 98\% of tokens in a text to comprehend it.
A follow-up study by \citet{Komori2004-es} estimates the percentage to be 96\% for Japanese learners to comprehend text.
Acquiring vocabulary to reach such levels, in turn, is a lengthy and challenging task for learners.
This opens up opportunities for assistive applications, such as simplification or annotation of complex words.
The first step necessary for such applications is to predict the complexity of the words.
The task of \textbf{lexical complexity prediction (LCP)} is defined as predicting how difficult to comprehend words or phrases in a text are on a continuous scale \citep{Shardlow2020-rd}. This differentiates LCP from complex word identification (CWI), i.e., binary classification of complex words \citep{Yimam2018-rw}.
As complexity is naturally perceived as continuous, a continuous scale used in LCP allows to represent it without loss of information.

The LCP research so far has been limited to English, for which two LCP datasets have been constructed \citep{Shardlow2020-rd, Shardlow2022-predicting}, and no such dataset has been created for Japanese. 
Meanwhile, there are a number of features specific to the Japanese language that could affect lexical complexity, and their effects have yet to be studied. For example, the Chinese characters, which are used extensively in Japanese, lower text readability \citep{Tateisi1988-yz}.

Previous studies on Japanese lexical complexity used pedagogical word lists to estimate complexity level. 
\citet{Nishihara2020-oh} modeled lexical complexity of words based on the Japanese Educational Vocabulary List \citep{Sunakawa2012-dm}.
The word list assigns a degree of difficulty to each item, based on the subjective judgment of Japanese language teachers, not learners themselves, and does not consider the learners' L1 background.

In light of this, we present JaLeCoN\footnote{
    JaLeCoN is available at \url{https://github.com/naist-nlp/jalecon}.
}, Dataset of \textbf{Ja}panese \textbf{Le}xical \textbf{Co}mplexity for \textbf{N}on-Native Readers.
Our dataset has the following key features:

\begin{asparaenum}[(1)]
    \item Complexity scores for single words as well as multi-word expressions (MWEs);
    \item Separate complexity scores from Chinese/Korean annotators and others, addressing the considerable advantage of the former in Japanese reading comprehension.
\end{asparaenum}

Our analysis reveals that the non-Chinese/Ko\-rean annotators perceive words of Chinese origin or containing Chinese characters as especially complex.
In the baseline experiment, we investigate the effectiveness of a BERT-based system in the Japanese LCP task, and how it varies according to the word complexity and L1 background.


\section{Task Setting}
\label{sec:task-setting}

Since Japanese has no explicit word boundaries, word segmentation is the first prerequisite for LCP. We use short unit words (\textbf{SUWs}) as the basic word unit, combining them into longer word units in the case of multi-word expressions (\textbf{MWEs}):

\begin{figure*}[t]
    \centering
    \footnotesize
	\newlength{\wMigi}
	\setlength{\wMigi}{\widthof{right.shoulder}+\widthof{上がり}+\tabcolsep}
	\newlength{\wTeiru}
	\setlength{\wTeiru}{\widthof{GER}+\widthof{be-PRS}+\tabcolsep}
	\newlength{\wNi}
	\setlength{\wNi}{\widthof{ADV}+\tabcolsep+1pt}
	\newlength{\wFue}
	\setlength{\wFue}{\widthof{increase}+\tabcolsep+2pt}
    \normalsize
	\newcommand{\MWE}{\hfill{}\raisebox{3pt}{\textcolor{HistogramOrange}{\tiny{\textsf{MWE}}}}}
	\newcommand{\MWEbox}[3]{\setlength{\fboxsep}{2pt}\fcolorbox{HistogramOrange}{white}{\parbox[t]{#1}{#2\MWE\\#3}}}
	\newcommand{\SUW}{\hfill{}\raisebox{3pt}{\textcolor{HistogramBlue}{\tiny{\textsf{SUW}}}}}
	\newcommand{\SUWbox}[3]{\setlength{\fboxsep}{2pt}\fcolorbox{HistogramBlue}{white}{\parbox[t]{#1}{#2\SUW\\#3}}}
    \begin{tabular}{l<{\hspace{1pt}}llllll}
\multirow[t]{2}{*}{\textbf{SUWs}} &
	\footnotesize 右肩 &
	\footnotesize 上がり &
	\footnotesize に & 
	\footnotesize 増え &
	\footnotesize て &
	\footnotesize いる\\[-0.5pt]
\footnotesize  &
	\footnotesize right.shoulder &
	\footnotesize rise &
	\footnotesize ADV &
	\footnotesize increase &
	\footnotesize GER &
	\footnotesize be-PRS\\[7pt]
\textbf{Words} &
	\multicolumn{2}{l}{%
		\MWEbox{\wMigi}{右肩上がり}{steady.rise}%
		} &
	\SUWbox{\wNi}{に}{ADV\phantom{j}} &
	\SUWbox{\wFue}{増え}{increase\phantom{j}} &
	\multicolumn{2}{l}{%
		\MWEbox{\wTeiru}{ている}{PRG-PRS\phantom{j}}%
		}\\[20pt]

& \multicolumn{6}{c}{``is steadily increasing''}\\[-2pt]
    \end{tabular}
    \caption{Example of text segmented as SUWs and as words (either \setlength{\fboxsep}{1pt}\fcolorbox{HistogramBlue}{white}{SUW} or \setlength{\fboxsep}{1pt}\fcolorbox{HistogramOrange}{white}{MWE}). Semantically opaque sequences are chunked into MWEs. Abbreviations in glosses: ADVerbializer, GERund, PReSent, PRoGressive.}
    \label{tab:example-words}
\end{figure*}

\begin{asparadesc}
    \item[SUW:] SUWs consist of one or two smallest lexical units \citep{ogura2011},
    and are commonly used for segmentation of Japanese.
    \item[MWE:] We understand MWEs as multi-\textit{SUW} expressions that are semantically opaque or institutionalized (see \cref{sec:mwe-category}) and consequently may have higher complexity than their components.
    We identify MWEs either using long unit word (LUW)\footnote{
        The LUW is defined as a syntactic word by \citet{omura-etal-2021-word}.
    } segmentation, or manually (see \cref{sec:construction}).
\end{asparadesc}

Consequently, a \textbf{word}, can be either an SUW or an MWE (see \cref{tab:example-words} for examples).

A \textbf{complexity score} represents perceived complexity based on the annotators' judgment on a scale from 0 (least complex) to 1 (most complex). We exclude proper nouns from our target because their complexity is influenced by factors unrelated to reading proficiency or vocabulary knowledge.\footnote{
    Sequences containing segmentation errors are also excluded (see \cref{sec:excluded}).
}

We annotate the words in an \textbf{in-context dense} setting. 
In-context here means including both intra-sentence and extra-sentence context of each word.
Context is important for lexical complexity for two reasons (\citealp{Gooding2019-ki}; \citealp{shardlow-etal-2021-semeval}):
(1) As polysemous words can have different complexity levels for each sense, context is necessary to differentiate between possible meanings of these words. 
(2) Presenting a word without context could increase its complexity. In particular, the recognition of abstract words relies on context \citep{Schwanenflugel1988}.
\textbf{Dense} means annotating each word of the text with a complexity label, instead of annotating one specific word in each sentence \citep{Shardlow2022-predicting}. 
We adopt the dense setting to avoid any bias that could arise from targeting specific words.


\section{Construction of JaLeCoN}
\label{sec:construction}

\begin{table*}[t]
\centering
    \begin{tabular}{lccccccc}
        \toprule
        &  &  &  & \multicolumn{2}{c}{CK} & \multicolumn{2}{c}{Non-CK} \\
        \cmidrule(lr){5-6}\cmidrule(lr){7-8}
        Genre & Sentences & Words & MWE Ratio 
        & All Words & MWEs & All Words & MWEs \\
        \midrule
        News & 400 & 10,256 & \phantom{0}7.9\% & .009 & .020 & .024 & .072 \\
        Government & 200 & \phantom{0}7,964 & 14.4\% & .005 & .009 & .028 & .047 \\
        \bottomrule
    \end{tabular}
    \caption{Statistics of JaLeCoN. The CK and Non-CK columns show the mean complexity scores by L1 group.}
    \label{tab:jalecon-stats}
\end{table*}

In order to include both written and spoken language and a variety of vocabulary, we sourced texts from two different genres:

\begin{asparadesc}
     \item[News] comes from the Japanese-English data of the WMT22 General Machine Translation Task \citep{Kocmi2022-wg}. It contains a variety of news texts written for the general Japanese reader. 
    \item[Government] is composed of press conference transcripts from Japanese ministries or agencies.\footnote{
        The transcripts were retrieved from the websites of five organizations: JMA, JTA, MOJ, MOFA, and MLHW.
    }
\end{asparadesc}

The whole dataset is composed of sequences of sentences constituting either the beginning of an article (News) or a question-answer pair (Government). 
We restricted the length of the sequences to at least 6 and at most 11 sentences to obtain similar amounts of text, and presented each sequence as a whole for annotation.

\subsection{Word Segmentation}
\label{sec:segmentation}

We used Comainu 0.80\footnote{\url{https://github.com/skozawa/Comainu}} \citep{Kozawa2014-po} to perform two-level segmentation. The low-level SUW segmentation was done using MeCab \citep{Kudo2004-qd}, a Japanese morphological analyzer, and the UniDic 2.3.0 \citep{Den2007-pf} dictionary. 
At the second level, Comainu chunked the SUWs into LUWs.
Based on the two segmentations, we segmented the text into words as follows:
\begin{asparaenum}[(1)]

\item If an LUW is a noun, we use the constituting SUWs as words. Transparent noun compounds are ubiquitous in Japanese (e.g., 
次期\split 気象\split 衛星\footnote{%
The vertical bars denote boundaries between SUWs.
} ``next meteorological satellite''), and we do not consider them MWEs.

\item If an LUW is not a noun, we use the LUW as a word. Such an LUW may be a single SUW, or a sequence of SUWs, which we consider an MWE. Such MWEs most importantly include functional words, such as compound particles (e.g., に\split つい\split て ``about'') and auxiliary verbs (e.g., なけれ\split ば\split なら\split ない ``have to'').
\end{asparaenum}

We also identified other MWEs manually, as explained in \cref{sec:mwe}.

\subsection{Complexity Annotation}
\label{sec:complexity}

To capture the lexical complexity for a non-native Japanese reader with intermediate or advanced reading ability, we recruited 15 annotators per sentence with Japanese reading proficiency ranging from CEFR (Common European Framework of Reference for Languages) level B1 to C2. 
We required at least intermediate proficiency, as it has been shown that complexity judgments made by intermediate or advanced learners can be used to adequately predict the needs of beginners but not vice versa \citep{gooding-etal-2021-word}. The proficiency levels were self-reported (see \cref{sec:annotator} for details). 
We used the annotations made by 14 of them, after removing one outlier, whose annotations had over 70\% higher mean than those of any other annotator, clearly not corresponding to the reported reading proficiency.

Approximately half of the annotators we recruited have a Chinese/Korean L1 background (CK).\footnote{
    On average, the CK annotators reported higher Japanese reading proficiency than the non-CK (see \cref{sec:annotator}).
} 
CK learners have a considerable advantage in comprehension of words of Chinese origin, which also form a large part of Chinese and Korean vocabulary \citep{Koda1989-ia}. 

The annotators were asked to assign one of the following labels to each span if they find it complex: 3 (Very Difficult), 2 (Difficult), or 1 (Not Easy); otherwise the annotators were to leave the span unlabeled and we interpreted it as 0 (Easy).\footnote{
    See \cref{sec:complexity-labels} for detailed definitions of each label.
}
Annotators could label a span of any length if it was complex as a whole, but were asked to create as short a span as possible.
To calculate the average, the labels were converted to numerical values as follows: 3\,→\,1, 2\,→\,0.67, 1\,→\,0.33, 0\,→\,0.
The averaging hinges on the assumption that the labels have an equal distance between them. We always presented the labels together with the values 0 to 3 to reinforce the perception of equal distance.

\subsection{MWE Annotation}
\label{sec:mwe}

In parallel with the complexity annotation, we annotated MWEs not identified by LUW segmentation (see \cref{sec:segmentation}). Given the absence of an MWE detector for Japanese of sufficient quality, 
the annotation was performed manually by a native Japanese speaker and a non-native speaker with a degree in the Japanese language. 
The expression categories we consider MWEs are described in \cref{sec:mwe-category}.

\subsection{Complexity Scoring}

Using annotations from the previous steps, we assigned complexity scores to words according to the following rules:
\begin{asparaenum}[(1)]
\item If a span contains one or more words, each word receives the complexity value of the span.
\item If an MWE (manually annotated according to \cref{sec:mwe})  overlaps with or contains multiple spans, the MWE receives the maximum of the complexity values of the spans.
\end{asparaenum}

Finally, for each word, we calculated the complexity score for each L1 group as the average of the individual values from the annotators in that group.


\section{Statistics and Analysis}
\label{sec:statistics}

Overall statistics for both genres and L1 groups are shown in \cref{tab:jalecon-stats}.\footnote{
    See \cref{sec:score-distribution} for the complexity scores and annotation distributions of several words in the non-CK group.
} MWEs have higher mean complexity than single words for both L1 groups and are more frequent in the Government genre. There is a tendency towards perceiving higher complexity in the non-CK group, which corresponds to slightly lower average Japanese proficiency of the non-CK annotators (see \cref{sec:annotator}).

We measured inter-annotator agreement (IAA) using Krippendorf's $\alpha$ for interval values \citep{krippendorff_bivariate_1970}. The IAA is 0.32 in the CK group, and 0.31 in the non-CK group, while it would be 0.19 if we merged the groups. As lexical complexity is highly subjective \citep{gooding-etal-2021-word}, the low agreement does not imply low reliability, but it indicates that perception of complexity is more alike within the L1 groups than across all annotators.

\begin{figure}[t]
    \centering
    \includegraphics[width=1.0\linewidth]{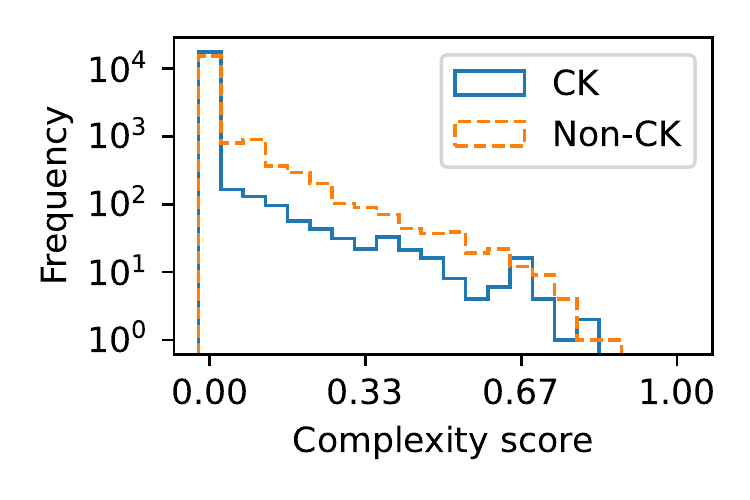}
    \vspace{-2\baselineskip} 
    \caption{Histogram of complexity scores by L1 group.}
    \label{fig:distribution}
\end{figure}

The complexity score distribution in each L1 group is shown in \cref{fig:distribution}.
No words achieved a score greater than 0.81 and 0.86 in the CK and non-CK groups, respectively, which reflects that words are rarely labeled as Difficult or Very Difficult by all annotators in a group.

\begin{table}[t]
\centering
    \begin{tabular}{
            >{\hspace{-2pt}}l
            <{\hspace{-1pt}}>{\hspace{-2pt}}c
            <{\hspace{-2pt}}>{\hspace{-2pt}}c
            <{\hspace{-1pt}}>{\hspace{-2pt}}c
            <{\hspace{-2pt}}>{\hspace{-2pt}}c
            <{\hspace{-1pt}}>{\hspace{-2pt}}c
            <{\hspace{-2pt}}>{\hspace{-2pt}}c
        }
        \toprule
        & \multicolumn{2}{c}{Japanese} & \multicolumn{2}{c}{Chinese} & \multicolumn{2}{c}{Other} \\
        \cmidrule(lr){2-3}\cmidrule(lr){4-5}\cmidrule(lr){6-7}
        & All & CC & All & CC & All & CC \\
        \midrule
        CK & .003 & \small{.009} & .004 & \small{.004} & .071 & \small{.000} \\
        Non-CK & .010 & \small{.032} & .062 & \small{.072} & .007 & \small{.143} \\
        \midrule
        Frequency & 52\% & \small{10\%} & 26\% & \small{22\%} & \phantom{0}4\% & \phantom{0}\small{0\%} \\
        \bottomrule
    \end{tabular}
\caption{Mean complexity (by L1 group) and frequency, according to (1) word origin: Japanese (\textit{wago}), Chinese/Sino-Japanese (\textit{kango}), and Other (\textit{gairaigo}, borrowings from languages other than Chinese), and (2) whether the words contain Chinese characters only (denoted by CC). The origin was classified using MeCab and Comainu (see \cref{sec:segmentation}), excluding words of mixed or unknown origin.
}
\label{tab:l1-by-word-type}
\end{table}

In addition to the aforementioned difference in proficiency, there is also a clear difference in how the two L1 groups perceive complexity of words based on their origin and whether they contain Chinese characters\footnote{
    Japanese vocabulary consists of words of Japanese origin, Chinese (Sino-Japanese) origin, and foreign words from other languages (\textit{gairaigo}). The first two categories can be written using Chinese characters (\textit{kanji}), Japanese syllabary (\textit{kana}), or a combination thereof, while other foreign words are usually written in syllabary only. (See \cref{sec:ex-origin} for examples.)}%
, as analyzed in \cref{tab:l1-by-word-type}. For the CK group, the mean complexity of words of Japanese and Chinese origin was similar. For the non-CK group, however, words of Chinese origin were markedly more complex (0.062) than words of Japanese origin (0.010), and both categories of words were more complex when they contained Chinese characters.\footnote{
    The opposite tendency for \textit{gairaigo} (foreign words mostly from English) to be perceived as more complex in the CK group coincides with lower English proficiency among annotators in this group (see \cref{sec:annotator}), and therefore should not be explained by their L1 background.}


\section{Experiments}
\label{sec:experiments}

The newly created dataset can be used to evaluate performance of LCP for non-native Japanese readers of different L1 backgrounds (CK and non-CK). We developed a baseline system based on a fine-tuned BERT \citep{devlin-etal-2019-bert} model, and evaluated it using cross-validation.
We fine-tuned a Japanese pre-trained BERT model released by Tohoku University, namely the base model for UniDic Lite segmentation\footnote{Available from \url{https://huggingface.co/cl-tohoku/bert-base-japanese-v2}.}.

For each word $w$ in our dataset and the sentence $s$ that contains it at token indices $i$ to $j-1$, we construct an input sequence
$(\texttt{[CLS]},$
$s_{0}^{i-1},$
$\texttt{<Unused1>},$
$w,$
$\texttt{<Unused2>},$
$s_j^{\left|s\right|-1},$
$\texttt{[SEP]},$
$w,$
$\texttt{[SEP]})$.
The target word occurs first delimited by unused tokens (\texttt{<Unused$n$>})
in the sentence context, and then on its own following the first $\texttt{[SEP]}$ token.\footnote{
    Due to a different segmentation (version of UniDic) used by Tohoku BERT and our dataset, we have to enforce segmentation at the word's boundaries using spaces.
}
To predict the complexity score, we feed the final hidden representation of the \texttt{[CLS]} token into a linear layer with a single output. A similar fine-tuning approach, but without the special tokens, was used for English LCP by \citet{taya-etal-2021-ochadai}, achieving one of the highest $R^2$ values in the single-word subtask of SemEval-2021 Task~1 \citep{shardlow-etal-2021-semeval}.

We fine-tune and evaluate models for CK and non-CK complexity separately.
See \cref{sec:experimental-setting} for the hyperparameters and cross-validation scheme.

\begin{table*}[t]
    \centering
    \begin{tabular}{lccccc} 
        \toprule
         & \multicolumn{4}{c}{MAE by Gold Complexity Score Tier} \\ 
         \cmidrule(lr){2-5}
         & \setulcolor{Zero}\ul{Zero} & \setulcolor{Easy}\ul{Easy $> 0$} & \setulcolor{NotEasy}\ul{Not Easy} & \setulcolor{Difficult}\ul{(Very) Difficult} & $R^2$ \\ 
         \midrule
         CK & 0.0034 & 0.0676 & 0.1913 & 0.2954 & 0.4351 \\
         Non-CK & 0.0066 & 0.0510 & 0.1169 & 0.2932 & 0.6142 \\
         \bottomrule
    \end{tabular}
\caption{Results of the fine-tuned BERT model by L1 group (means over 5 cross-validation folds).}
\label{tab:results}
\end{table*}

\begin{table*}[!ht]
    \centering
    \begin{tabular}{lcccc}
        \toprule
        & \setulcolor{Zero}\ul{Zero} & \setulcolor{Easy}\ul{Easy $> 0$} & \setulcolor{NotEasy}\ul{Not Easy} & \setulcolor{Difficult}\ul{(Very) Difficult} \\
        \midrule
        CK & 17,563 & \phantom{0,}393 & 223 & \phantom{0}41 \\
        Non-CK & 15,209 & 2,067 & 837 & 107 \\
        \bottomrule
    \end{tabular}
    \caption{Word counts in the whole dataset by L1 group and MAE tier.}
    \label{tab:word-counts}
\end{table*}

\begin{figure*}[!ht]
    \centering
    \includegraphics[width=0.8\linewidth]{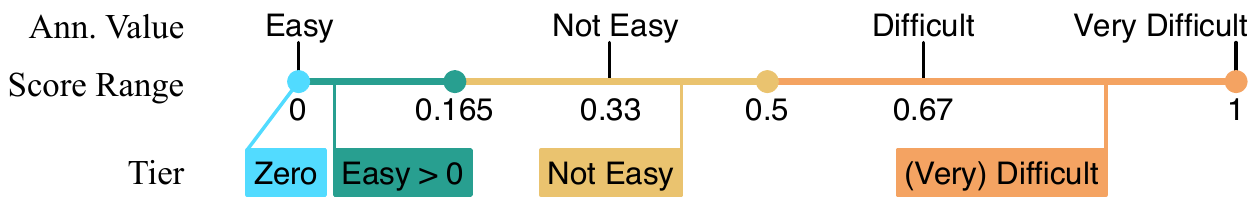}
    \caption{Illustrated score ranges of the MAE tiers: $\{0\}$ for Zero, $(0, 0.165]$ for Easy $> 0$, $(0.165, 0.5]$ for Not Easy, and $(0.5, 1]$ for (Very) Difficult.}
    \label{fig:mae-tiers}
\end{figure*}




The results are reported in \cref{tab:results}. 
In addition to $R^2$ (coefficient of determination)\footnote{
Compared to correlation coefficients, $R^2$ is more appropriate for LCP, since it also captures deviations in mean and variance. Compared to MAE or MSE, it is easier to interpret, as $R^2 = 0$ corresponds to the mean regressor, while $R^2 = 1$ corresponds to a perfect model.
}, we report mean average error (MAE) by complexity score tiers to draw the full picture of the models' performance at different complexity levels.
The score ranges of the tiers are centered at annotation values as illustrated in \cref{fig:mae-tiers}. We handle zero as a special tier, and merge Very Difficult with Difficult due to a low number of words.

The fine-tuned BERT model for CK and Non-CK achieves $R^2$ of 0.4351 and 0.6142, respectively.
For both L1 groups, the MAE value increases markedly in each successive complexity tier, as the number of training examples (shown in \cref{tab:word-counts}) diminishes.
Similarly, the CK model achieves lower error than non-CK only in tier Zero, where it has more examples available than the non-CK model. This suggests that the scarcity of words with complexity above zero is a factor contributing to worse performance on CK data, as measured by~$R^2$.

\section{Conclusion}

In this paper, we presented the first dataset for Japanese LCP. It provides separate complexity scores based on the CK/non-CK distinction of annotators' L1 background. 
Our analysis corroborates our conjecture that special consideration of L1 background is useful for the Japanese LCP task in particular.
We believe it could benefit LCP in other languages as well.

In the baseline experiment, we demonstrated the efficacy of our BERT-based system for both CK and non-CK readers.
Even after separating CK and non-CK annotators, however, notable inter-annotator disagreement remains within these groups. 
Therefore personalized systems analogous to \citet{gooding-tragut-2022-one} could improve on our system. Future research should study this possibility, analyzing both its costs and benefits.

Models trained on JaLeCoN can be used as part of a lexical simplification pipeline for Japanese, both to identify complex words and to rank candidate simplifications. 
JaLeCoN itself can be further used as a basis for a lexical simplification dataset targeting words actually perceived as complex, similar to TSAR-ST datasets for English and Spanish \citep{Stajner2022-nb}. 

\section*{Limitations}

Our task setting and baseline system requires that the input is already segmented into words including MWEs. 
The MWE identification step in the construction process of our dataset involved time-consuming manual annotation.
Building a high-quality system that fully automates the process is an issue for future work.
Our dataset can be used to evaluate such a Japanese MWE identification system.

Additionally, as shown in \cref{sec:experiments}, our baseline model performed relatively poorly in the higher complexity tiers.
This is an effect of the dense annotation setting; it results in uneven distributions of complexity as shown in \cref{fig:distribution}, where easy words greatly outnumber difficult words.
One possible solution would be creating another LCP dataset using sparse annotation, where target words are selected using frequency bands so that the words are distributed across a wide range of frequency \citep{Shardlow2022-predicting}.
Our data could provide insights as to what kind of words should be targeted by sparse annotation for such a dataset.


\section*{Acknowledgments}
We would like to express our gratitude to Justin Vasselli and the anonymous reviewers for their insightful feedback.
This work was supported by JSPS KAKENHI grant number JP19K20351 and NAIST Foundation.



\bibliography{anthology,custom}
\bibliographystyle{acl_natbib}


\newpage

\appendix
\onecolumn

\section{Annotators}
\label{sec:annotator}

\begin{table}[ht]
\centering
    \begin{tabular}{lcccc}
        \toprule
        & \multicolumn{2}{c}{Japanese} & \multicolumn{2}{c}{English} \\
        \cmidrule(lr){2-3} \cmidrule(lr){4-5} 
        & B1/B2 & C1/C2 & B1/B2 & C1/C2 \\ 
        \midrule
        CK & 4 & 3 & 7 & 0 \\
        Non-CK & 6 & 1 & 2–3 & 4–5\\ 
        \bottomrule
    \end{tabular}
\caption{Annotator counts per sentence in each L1 group, by Japanese and English reading proficiency category. The proficiency levels were determined by self-reports with reference to an assessment grid either in Japanese\footnotemark{} or in English.\footnotemark{} Overall, our CK annotators are better at Japanese reading and poorer at English reading than the non-CK.}
\end{table}

\begin{table}[ht]
\centering
    \begin{tabular}{lllll}
        \toprule
        CK & Chinese: 6, &Korean: 1 \\
        Non-CK & English: 2–3, &Thai: 2–3, &Indonesian: 1, &Lao: 1 \\
        \bottomrule
    \end{tabular}
\caption{Annotator counts per sentence of each L1 group.}
\end{table}

\addtocounter{footnote}{-1}\footnotetext{
    \url{https://jfstandard.jp/pdf/self_assessment_jp.pdf}
}
\addtocounter{footnote}{+1}\footnotetext{
    \url{https://rm.coe.int/CoERMPublicCommonSearchServices/DisplayDCTMContent?documentId=090000168045bb52}
}


\section{Complexity labels}
\label{sec:complexity-labels}

\begin{table}[ht]
\centering
    \begin{tabular}{l p{0.72\linewidth}}
        \toprule
        3 (Very Difficult): & You hardly understand its meaning in the context. \\
        2 (Difficult): & You can infer its meaning, but you are not confident. \\
        1 (Not Easy): & You understand its meaning with confidence, but it is quite difficult among the expressions you can understand. \\
        0 (Easy): & None of the above. \\
        \bottomrule
    \end{tabular}
    \caption{Complexity labels. An annotator can label spans with complexity 3, 2, 1, or 0.}
    \label{tab:labels}
\end{table}


\newpage

\section{MWE Categories}
\label{sec:mwe-category}

\begin{table}[ht]
\centering
    \begin{tabular}{p{0.16\linewidth}p{0.56\linewidth}>{\raggedright\arraybackslash}p{0.2\linewidth}}
        \toprule
        Category & Description & Example \\ 
        \midrule
        Lexicalized \mbox{expressions} & Non-compositional expressions whose meaning as a whole cannot be completely inferred by the meaning of their components. & 使い\split 勝手 \mbox{(ease of use)} \\
        \midrule

        Institutionalized expressions & Compositional expressions whose components cannot be replaced without distorting the meaning of the whole expression or violating the language conventions. & 感染\split 症 \mbox{(infectious disease)} \\
        \midrule
        
        Functional \mbox{expressions} & Expressions that behave like single function words. & に\split つき\split まし\split て \mbox{(as for)} \\
        \bottomrule
    \end{tabular}
\caption{Categories we regard as MWEs. See \citet{Kochmar2020-ep} for lexicalized and institutionalized expressions, and \citet{Matsuyoshi2007-to} for functional expressions. The vertical bars in the examples denote boundaries between SUWs.}
\end{table}


\section{Excluded categories}
\label{sec:excluded}

\begin{table}[ht]
\centering
    \begin{tabular}{p{0.16\linewidth}p{0.56\linewidth}>{\raggedright\arraybackslash}p{0.2\linewidth}}
        \toprule
        Category & Identification Approach & Example \\ 
        \midrule
        Proper nouns & Proper nouns are first identified by MeCab. We also manually annotate proper noun phrases. & \mbox{関東\split 大\split 震災} \mbox{(The Great Kantō} Earthquake) \\
        \midrule
        Segmentation \mbox{errors} & We manually annotate sequences with segmentation errors. & も\split や (mist) \\ 
        \bottomrule
    \end{tabular}
\caption{Categories of words or spans we exclude from our target. The vertical bars in the examples denote boundaries between SUWs. The correct segmentation for も\split や is もや.}
\end{table}


\newpage

\section{Distributions of Annotation}
\label{sec:score-distribution}

\begin{figure*}[ht]
    \centering
    \includegraphics[width=1.0\linewidth]{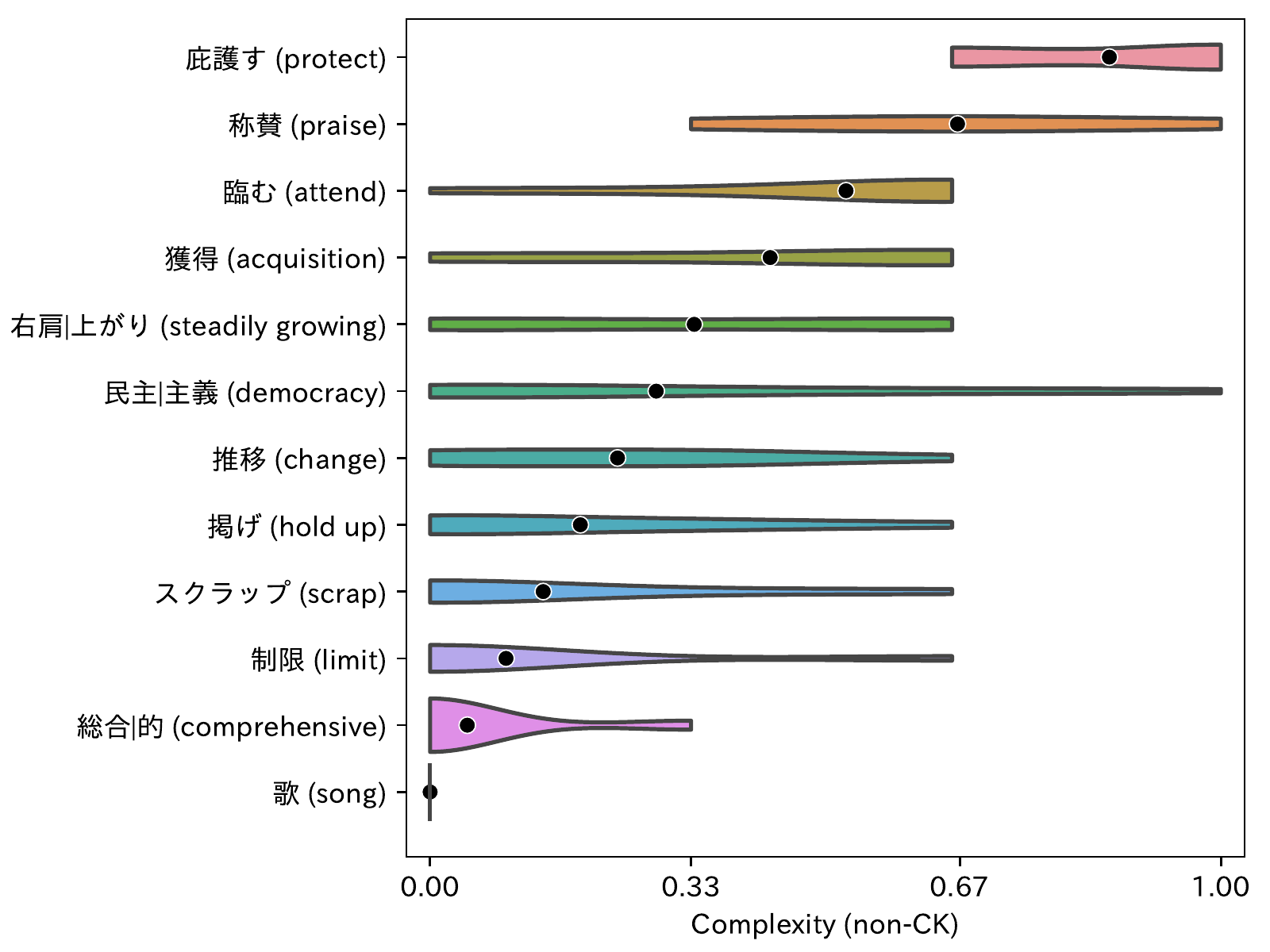}
    \caption{Violin plot showing annotation distributions of several words with dot markers showing the complexity scores, both for non-CK annotators. Words are shown in their surface forms; the vertical bars in them denote boundaries between SUWs.}
    \label{fig:distribution-violin}
\end{figure*}

\vfill
\section{Examples of Words by Origin}
\label{sec:ex-origin}

\vspace{\baselineskip}
\noindent%
\begin{minipage}{\textwidth}%
\captionsetup{type=table}%
\newlength{\wHP}
\setlength{\wHP}{\widthof{ホーム\split ページ\enspace (home page <Eng)}-0.64pt}
\centering%
    \begin{tabular}{l >{\raggedright}p{\wHP} l}
        \toprule
        \multirow{2}{*}{Origin} & \multicolumn{2}{c}{Containing Chinese characters (\textit{kanji})?}\\
        \cmidrule(lr){2-3} 
        & \multicolumn{1}{c}{Yes} & \multicolumn{1}{c}{No}\\
        \midrule
        Japanese & 
        歌\enspace (song) &
        けれど\split も\enspace (although) \\ 
        (\textit{wago}) & 
        臨む\enspace (attend) &
        ふさわしい\enspace (suitable) \\ 
        \midrule
        Chinese/Sino-Japanese &
        今回\enspace (this time) &
        よう\enspace (it seems \textdagger{}様)\\
        (\textit{kango}) &
        民主\split 主義\enspace (democracy) &
        もちろん\enspace (of course \textdagger{}勿論)\\
        \midrule
        Other &
        旦那\enspace (husband <Skt) &
        スクラップ\enspace (scrap <Eng)\\
        (\textit{gairaigo}) & & ホーム\split ページ\enspace (home page <Eng)
        \\
        \bottomrule
    \end{tabular}
\captionof{table}{Examples of words in JaLeCoN categorized by word origin and whether they contain Chinese characters. \textdagger{} marks a variant of the word written using Chinese characters documenting the Sino-Japanese origin; < marks the word's origin (Sanskrit or English). The vertical bars in the examples denote boundaries between SUWs. All categories except Other (\textit{gairaigo}) written using Chinese characters are relatively common.
}
\end{minipage}


\newpage

\section{Experimental Setting}
\label{sec:experimental-setting}

\begin{table}[ht]
\centering
    \begin{tabular}{lr}
        \toprule
        Optimizer & Adam ($\beta_1 = 0.9$, $\beta_2 = 0.999$) \\ 
        \ -- learning rate & 5e-5 \\
        \ \phantom{--} -- schedule & no warm-up, linear decay \\ 
        \ -- L2 weight decay & 0.01 \\
        Epochs & 5 \\
        Loss function & Mean squared error \\
        Dropout & 0.1 \\
        Batch size & 16 \\
        Weight initialization & $\mathcal{N}(\mu=0,\,\sigma=0.02)$ truncated to $\pm 2\sigma$ \\
        Bias initialization & 0 \\
        Gradient L2 norm clipping & 2 \\
        \bottomrule
    \end{tabular}
\caption{Hyperparameters used for fine-tuning the BERT model. We have chosen the combination of learning rate (from 8e-6, 5e-5, 3e-5, and 2e-5), warm-up (from no warm-up and 10\% steps), and the number of epochs (from 1 to 5) achieving the highest mean $R^2$ in a nested 4-fold cross-validation on the training data of the first outer cross-validation split. The optimal combination was identical for CK and non-CK complexity.}
\label{tab:hyperparam}
\end{table}

\begin{table}[ht]
\centering
    \begin{tabular}{lr}
        \toprule
        Folds & 5 \\ 
        Stratification & by genre (News and Government) \\
        Grouping & by sequence of sentences (see \cref{sec:construction}) \\ 
        \bottomrule
    \end{tabular}
\caption{Cross-validation scheme used for fine-tuning and evaluation of the BERT model.}
\label{tab:cv}
\end{table}


\end{document}